\documentclass[10pt,twocolumn,letterpaper]{article}

\usepackage{cvpr}              
\usepackage{makecell}
\usepackage{multirow}
\usepackage{booktabs}
\usepackage[table]{xcolor}
\usepackage{pifont}
\definecolor{cvprblue}{rgb}{0.21,0.49,0.74}
\usepackage[pagebackref,breaklinks,colorlinks,citecolor=cvprblue]{hyperref}

\title{DiffVLA: Vision-Language Guided Diffusion Planning for Autonomous Driving}

\author{Anqing Jiang$^{1}$\footnotemark[1] \quad Yu Gao$^{1}$ \quad Zhigang Sun $^{1}$ \quad  Yiru Wang$^{1}$ \quad Jijun Wang$^{2}$  \quad Jinhao Chai$^{1,3}$ \\
Qian Cao$^{1,5}$ \quad Yuweng Heng$^{1}$ \quad Hao Jiang$^{1,4}$ \quad YunDa Dong$^{1,4}$ \quad Zongzheng Zhang$^{2}$ \quad Xianda Guo$^{2}$ \\ Hao Sun$^{1}$ \quad Hao Zhao$^{2}$ \\
 \quad  $^{1}$RIX, Bosch \quad 
$^{2}$AIR, Tsinghua University \quad
$^3$Shanghai University \quad
$^4$Shanghai Jiao Tong University \\
$^5$Southeast University\\
\vspace{-0.5cm}
\\Team: RB\\
{\tt\small \{anqing.jiang\}@cn.bosch.com}
}

\begin{document}
\maketitle
\renewcommand{\thefootnote}{\fnsymbol{footnote}}
\footnotetext[1]{Corresponding authors.}
\renewcommand{\thefootnote}{\arabic{footnote}}

\begin{abstract}

Research interest in end-to-end autonomous driving has surged owing to its fully differentiable design integrating modular tasks, i.e. perception, prediction and planing, which enables optimization in pursuit of the ultimate goal. Despite the great potential of the end-to-end paradigm, existing methods suffer from several aspects including expensive BEV (bird’s eye view) computation, action diversity, and sub-optimal decision in complex real-world scenarios. To address these challenges, we propose a novel hybrid sparse-dense diffusion policy, empowered by a Vision-Language Model (VLM), called Diff-VLA. We explore the sparse diffusion representation for efficient multi-modal driving behavior. Moreover, we rethink the effectiveness of VLM driving decision and improve the trajectory generation guidance through deep interaction across agent, map instances and VLM output. Our method shows superior performance in Autonomous Grand Challenge 2025 which contains challenging real and reactive synthetic scenarios. Our methods achieves 45.0 PDMS.

\end{abstract}
\section{Introduction}
End-to-end autonomous driving has emerged as a significant and rapidly growing research area. With the abundance of available human driving demonstrations, there is considerable potential to learn human-like driving policies from large-scale datasets.  Methods such as UniAD \cite{hu2023planning}, VAD \cite{jiang2023vad} take sensor data as input and regress a single-mode trajectory within one fully optimizable model. \cite{sun2024sparsedrive} further explores the sparse presentation and propose a symmetric sparse perception module and a parallel motion planner.However, these approaches overlook the intrinsic uncertainty and the multi-modal characteristics of driving behavior. Leveraging powerful diffusion concepts in generation, methods \cite{chitta2022transfuser,zheng2025diffusion} are capable of modeling multi-mode action distributions. Through proper denoising process,  \cite{liao2024diffusiondrive} further speeds up diffusion process via an anchored Gaussian distribution design. \cite{jiang2024senna, zhou2025opendrivevla} connects VLM to an end-to-end model to improve the trajectory planning accuracy.

Although existing methods are trained and evaluated on well-known benchmarks such as nuScenes, NAVSIMv1, nuPlan,
achieving robust performance in a close-loop manner and going beyond recorded states remains a open challenge. In this report, we revisit the concepts of sparsity, diffusion, VLM and propose a more comprehensive approach with proven performance in closed-loop assessment. Our framework consists of three key components: a VLM guidance module, a sparse-dense hybrid perception module, and a diffusion-based planning module.
\begin{enumerate}
        \item \textbf{VLA Guidance Module}: The module takes multi-view images as input and outputs both a trajectory and high-level driving commands. These commands are then combined with the external driving command (e.g. navigation instructions) to serve as input for the diffusion-based planning module.
        \item \textbf{Hybrid Perception Module}: Our hybrid perception model have two branch for different perception tasks. The dense perception branch constructs a dense Bird's Eye View (BEV) feature representation, which is fed into the planning module as a primary input. To improve the planner's understanding of obstacles and road structure, the sparse perception branch extracts information at the instance level (e.g., detected obstacles, lane boundaries, centerlines, stop lines, etc.) and propagates it to the planning module.
        \item \textbf{Diffusion-Based Planning Module}: We use truncated diffusion policy that utilizes multi-modal anchors as priors and employs a shortened diffusion schedule. To further improvement the diffusion model performance, we propose a hierarchical information encoding strategy to integrate heterogeneous inputs. 

\end{enumerate}
We train and evaluate our approach on navsim-v2, which  provides a comprehensive closed-loop assessment of robustness and generalization via introducing reactive background traffic participants and realistic synthetic multi-view camera images. Leveraging these techniques, our solution achieves 45.0 EPDMS on the private test set in navsim v2 competition.

\begin{figure*}[htp]
    \centering
    \includegraphics[width=\textwidth]{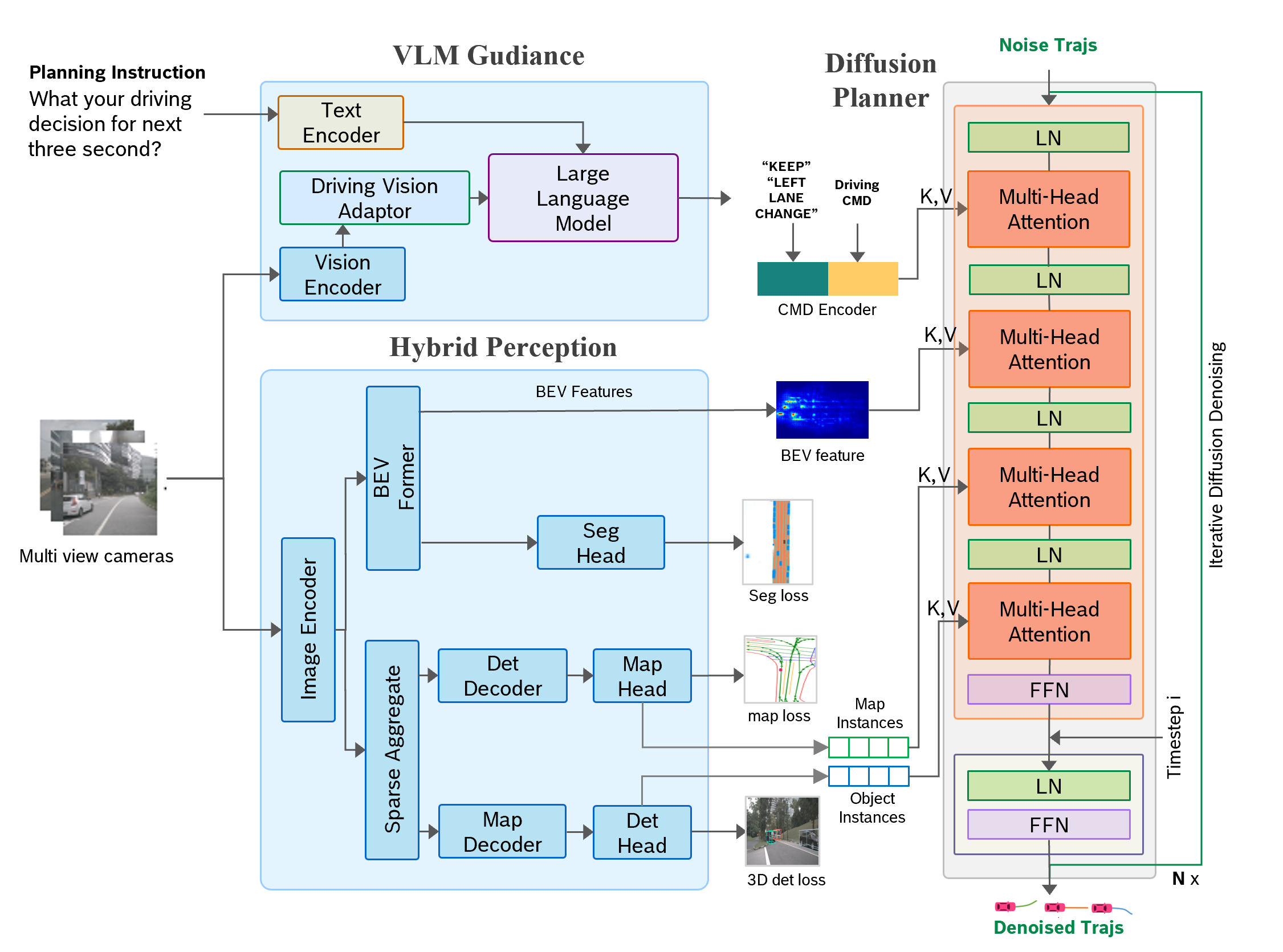}
    \caption{The architecture of our proposed perception enhanced diffusion VLA framework.}
    \label{fig:galaxy}
\end{figure*}

\section{Perception}

Agents and map information are critical for representing a traffic scene. Our design philosophy aims to maximize the representation of both explicit and implicit features.

In our current solution, we introduce two parallel perception modules: a sparse perception module and a dense perception module. The sparse perception module employs the sampling strategy from \cite{Lin2023Sparse4D} and \cite{jiang2025sparsemextunlockingpotentialsparse} for 3D object detection and online map generation. In contrast, the dense module utilizes the BEV feature projection method from \cite{huang2021bevdet} to produce a BEV feature space. The sparse module outputs 3D bounding boxes and map vectors, while the dense module generates BEV feature vectors. Both outputs are integrated into a subsequent trajectory head.

The purpose of these dual modules is to simultaneously leverage implicit features of agents and the environment alongside explicit object and map information, overcoming the limitations of using only projection-based or sampling-based methodologies for constructing the BEV feature space. Explicit 3D bounding boxes are formatted as standard object boxes, containing pose, size, heading angle, and velocity. Map vectors are represented as 20 map points per element. Both explicit object and map-related information are encoded using a MLP to generate explicit object and map embeddings.

For the implicit branch, the BEV grid size is set to $128 \times 128$, covering a perception range of $64 \times 64$ meters along the $x$ and $y$ directions in the ego coordinate space. We aggregate features from 30 agents and one ego vehicle to provide implicit guidance for the subsequent trajectory diffusion process. Additionally, the explicit outputs of objects and maps enable the planner to perform collision detection and drivable area checks, enhancing the planning process beyond feature-based trajectory selection alone.

The training process for the perception modules is divided into two stages. The sparse branch is trained using detection losses for 3D objects and map elements. Subsequently, the dense branch is trained alongside the trajectory head, following the completion of sparse branch training. All perception branches utilize the VoV-99 \cite{lee2020centermask} backbone.

\section{VLM}

To enable effective processing and fusion of multimodal information in autonomous driving scenarios, we propose the VLM Command Guidance module. This module is built upon the Senna-VLM framework~\cite{jiang2024senna}, which leverages a multi-image encoding strategy and multi-view prompting mechanisms to achieve efficient and comprehensive scene understanding.

The Senna-VLM architecture comprises four primary components: Vision Encoder, Driving Vision Adaptor, Text Encoder, and Large Language Model (LLM). The Vision Encoder processes multi-view image sequences $I \in \mathbb{R}^{N_{\mathrm{img}} \times H \times W \times 3}$ from Navsim~\cite{dauner2024navsim} as input and extracts image features. These features are further encoded and compressed by the Driving Vision Adapter, producing image tokens $E_{\mathrm{img}} \in \mathbb{R}^{N_{\mathrm{img}} \times M_{\mathrm{img}} \times C}$, where $N_{\mathrm{img}}$ is the number of images, $M_{\mathrm{img}}$ is the number of image tokens per image, $C$ is the feature dimension of the LLM, $H$ is the height of the image and $W$ is the width of the image. 

The Text Encoder encodes user instructions and navigation commands into text tokens $E_{\mathrm{txt}} \in \mathbb{R}^{M_{\mathrm{txt}} \times C}$, with $M_{\mathrm{txt}}$ signifying the number of text tokens. Both the image and text tokens are then fed into a Large Language Model (LLM), which generates high-level driving decisions.

In our implementation, the Vision Encoder adopts ViT-L/14 from CLIP~\cite{radford2021learning} and the LLM is Vicuna-v1.5-7B~\cite{zheng2023judging}. We adhere to the standard Senna-VLM configuration by processing images from all onboard camera sensors. 

The VLM Command Guidance module produces high-level planning decisions, which are decomposed into lateral control (e.g., lane changes and turning) and longitudinal control (e.g., acceleration and braking). These decisions are encoded via a one-hot encoding mechanism and subsequently integrated with external driving signals, such as navigation instructions. The resulting commands are then processed by a command encoder module, providing semantic guidance for the downstream diffusion-based planning process.

\section{Planning}

\vspace{-0.1cm}
We define a set of trajectory vocabularies $V=\left\{ v_i \right\}_{i=1}^{N}$ to discretize the action space of the ego vehicle, where $N$ represents the number of trajectory vocabularies. Each vocabulary $v_i$ consists of a set of waypoints $\tau=\left\{(x_t,y_t,\theta_t)\right\}{t=1}^{T_h}$, with a planning time horizon $T_h=4$. To construct the trajectory vocabularies, we sample all trajectories from the NavTrain split and apply k-means clustering to group them into discrete vocabularies.

Following the main idea of \cite{liao2024diffusiondrive}, we add Gaussian noise to $V$ to generate trajectory anchors $A=\left\{a_k\right\}_{k=1}^{N{\text{anchor}}}$, where $N_{\text{anchor}}=32$, and initiate the diffusion process. We utilize both explicit and implicit perception results from the sparse and dense BEV branches as explicit guidance $z_e$ from sparse BEV branch and implicit guidance $z_i$ from dense BEV branch, respectively, to enhance the guidance condition $z=\left\{z_i,z_e\right\}$ for the trajectory diffusion process. The diffusion model is formulated as:
\[
\left\{ \hat{s}_k, \hat{\tau}_k \right\}_{k=1}^{N_{anchor}} = f_{\theta}\left( \left\{ \tau_{k}^{i} \right\}_{k=1}^{N_{anchor}}, z \right)
\]
where $f_{\theta}$ represents the diffusion trajectory decoder, $\hat{\tau}_k$ denotes the predicted target waypoints, and $\hat{s}_k$ represents the classification score for the waypoints. The trajectory head is trained using a combination of trajectory classification and regression losses, following the same strategy as \cite{liao2024diffusiondrive} for selecting positive and negative samples.

\section{Post processing}


\subsection{Trajectory re-distribution}
We have observed that under high-speed conditions, the collision rate of predicted trajectories is slightly higher than in other scenarios. We believe this is caused by a deviation between the predicted and actual data distributions in terms of trajectory. Based on this assessment, we have implemented a 2\% deceleration along the y-axis for our trajectories. We found that this approach effectively prevents collisions without  significant differences were noted in other evaluation scores.
\vspace{-0.1cm}

\section{Experiments}

\subsection{Experiments Setup}
\vspace{-0.1cm}
\subsubsection{Multi-stage Training} The training of our solution is divided into two stages. In stage-1, we train the VLA gudiance model and hybrid perception module from scratch to learn the scene representation. In stage-2 VLA gudiance, hybrid perception module and parallel motion planner are trained together. Through extensive experimentation, we froze the VLA guidance model and the sparse perception model weights for superior training performance. The dense perception are trained togeher with no model weights frozen.

\subsubsection{Training Details}

\begin{table}[htbp]
\centering
\caption{Training hyperparameters}
\label{tab:hyperparameters}
\begin{tabular}{l|lccccc}
\hline
\textbf{Model}  & Training stage & Bs & Ep & Lr\\
\hline
\textbf{VLM}  & Stage-1 & 192 & 1 & 2e-5\\
\textbf{Sparse Per}  & Stage-1  & 256 & 100 & 1e-4\\
\textbf{Dense Per}  & Stage-1\&2  & 256 & 100 & 1e-4\\
\textbf{DiffPlan} & Stage-2  & 256 & 100 & 1e-4\\
\hline
\end{tabular}
\end{table}

Our model is trained in two stages. The first stage includes the sparse perception branch and the VLM module, which are trained separately. The sparse perception branch is trained in the first stage with a batch size of 128, a learning rate of $10^{-4}$, and a total of 100 epochs. The VLM is trained with a batch size of 192, a learning rate of $2\times10^{-5}$, and 1 epochs. The second stage involves training the dense perception branch and the trajectory head, using a batch size of 256, an initial learning rate of $10^{-4}$, and 100 epochs. Both stages employ the AdamW optimizer and a cosine learning rate decay policy. The training hyperparameters and stages are summarized in Table~\ref{tab:hyperparameters}.

\subsection{Experiments result}
 We present our proposed VLA architecture's results as shown in Table~\ref{tab:metrics_list}.
\begin{table}[htbp]
\centering
\caption{Results of proposed VLA architecture in NAVSIM2}
\label{tab:metrics_list}
\begin{tabular}{l|l}
\hline
\textbf{Metric Name}  & \textbf{Team: RB} \\
\hline
\textbf{extended\_pdm\_score\_combined}              & \textbf{45.007} \\
\hline
no\_at\_fault\_collisions\_stage\_one                & 95.7143\\
drivable\_area\_compliance\_stage\_one               & 99.2857\\
driving\_direction\_compliance\_stage\_one           & 100\\
traffic\_light\_compliance\_stage\_one               & 100\\
ego\_progress\_stage\_one                            & 85.9777\\
time\_to\_collision\_within\_bound\_stage\_one       & 96.4286\\
lane\_keeping\_stage\_one                            & 97.1429\\
history\_comfort\_stage\_one                         & 95\\
two\_frame\_extended\_comfort\_stage\_one            & 84.2857 \\
no\_at\_fault\_collisions\_stage\_two                & 81.2672\\
drivable\_area\_compliance\_stage\_two               & 88.8443\\
driving\_direction\_compliance\_stage\_two           & 94.6578\\
traffic\_light\_compliance\_stage\_two               & 99.0814\\
ego\_progress\_stage\_two                            & 86.0899 \\
time\_to\_collision\_within\_bound\_stage\_two       & 76.4597\\
lane\_keeping\_stage\_two                            & 59.8548\\
history\_comfort\_stage\_two                         & 98.6339\\
two\_frame\_extended\_comfort\_stage\_two            & 80.0481\\
\hline
\end{tabular}
\end{table}

\section{Limitation}
Due to time constraints, we train the VLM, sparse perception, dense perception and planning head separately. Combining all these modules for end-to-end training in a single stage remains unexplored, which may yield potential performance improvements.

\section{Conclusion}
In the paper, we propose a VLA that leverages the advantages of E2E autonomous driving, combined
with vlm gudiance, hybrid perception and diffusion planner. As a result, as shown in the last row of Tab.\ref{tab:metrics_list}, the proposed framework metric reached 45.0.



\bibliographystyle{IEEEtran}
\bibliography{main}


\end{document}